\documentclass{article}
\usepackage[a4paper]{geometry} % Adjust margins and paper size
\usepackage{makecell} % Include this in your preamble

\usepackage{url, hyperref}
\usepackage{cite}
\usepackage{amsmath,amssymb,amsfonts}
\usepackage{graphicx}
\usepackage{textcomp}
\usepackage{soul}
\usepackage{xcolor}
\sethlcolor{yellow}
\usepackage{multirow}
\usepackage{algorithm}
\usepackage{algpseudocode}
\usepackage{amsmath}

\begin{document}

    \title{QuIM-RAG: Advancing Retrieval-Augmented Generation with Inverted Question Matching for Enhanced QA Performance}

        \author{
        Binita Saha\textsuperscript{*}, Utsha Saha and Muhammad Zubair Malik\\
        \textit{\{binita.saha, utsha.saha, zubair.malik\}@ndsu.edu}\\
        North Dakota State University, Fargo, ND, USA\\
        \textsuperscript{*}\textit{Corresponding author: Binita Saha, binita.saha@ndsu.edu}
        \\\\
        \small{Draft article, final version DOI: \url{https://doi.org/10.1109/ACCESS.2024.3513155}} 
        \\
        \small{Published in \textit{IEEE ACCESS, volume-12}}
    }
    \maketitle
    \date{}
    
    \begin{abstract}

        This work presents a novel architecture for building Retrieval-Augmented Generation (RAG) systems to improve Question Answering (QA) tasks from a target corpus. Large Language Models (LLMs) have revolutionized the analyzing and generation of human-like text. These models rely on pre-trained data and lack real-time updates unless integrated with live data tools. RAG enhances LLMs by integrating online resources and databases to generate contextually appropriate responses. However, traditional RAG still encounters challenges like information dilution and hallucinations when handling vast amounts of data. Our approach addresses these challenges by converting corpora into a domain-specific dataset and RAG architecture is constructed to generate responses from the target document. We introduce QuIM-RAG (Question-to-question Inverted Index Matching), a novel approach for the retrieval mechanism in our system. This strategy generates potential questions from document chunks and matches these with user queries to identify the most relevant text chunks for generating accurate answers. We have implemented our RAG system on top of the open-source Meta-LLaMA3-8B-instruct model by Meta Inc. that is available on Hugging Face. We constructed a custom corpus of 500+ pages from a high-traffic website accessed thousands of times daily for answering complex questions, along with manually prepared ground truth QA for evaluation. We compared our approach with traditional RAG models using BERT-Score and RAGAS, state-of-the-art metrics for evaluating LLM applications. Our evaluation demonstrates that our approach outperforms traditional RAG architectures on both metrics.
    \end{abstract}


\newpage

    \section{Introduction}
    \label{sec:introduction}

   Large Language Models (LLMs) have revolutionized the field of Natural Language Processing (NLP) and fundamentally changed the way we used to interact with digital information. They have the capability to analyze and generate text that resembles human language, which has become integral across various sectors \cite{brown2020language}. However, when it comes to answering domain-specific questions where accuracy and reliability are key, the effectiveness of these models faces challenges due to knowledge limiation, contextual misinterpretation, task-specific variability, etc.  \cite{wang2023empower, kandpal2023large}.

    To address these limitations of adapting LLMs for specialized domains, two principal strategies have emerged: a) fine-tuning the models with domain-specific data, and b) augmenting them with external knowledge. Fine-tuning involves additional training of a pre-existing LLM using specialized data, which optimizes the model for particular tasks and significantly boosts its performance \cite{brown2020language}. Despite its effectiveness, fine-tuning is not without drawbacks; it is computationally demanding, costly, and runs the risk of forgetting critical information—where the model loses its ability to recall previously learned information. \cite{ovadia2023fine, kirkpatrick2017overcoming, goodfellow2013empirical, chen2020recall}. Moreover, the effectiveness of fine-tuning is dependent on the availability of extensive relevant data, which makes it less practical for domain-specific fields.

    Another approach to overcome the challenges of customizing LLMs for domain-specific tasks and enhancing precision in generating Q\&A responses is the use of Retrieval-Augmented-Generation (RAG) \cite{Saha2024}. The foundation of RAG is its capability to incorporate relevant information from external sources to ensure the generated responses are contextually appropriate \cite{lewis2020retrieval, neelakantan2022text}. This method allows LLMs to adapt to domain-specific tasks more flexibly, sidestepping the high costs and potential knowledge loss associated with fine-tuning. However, RAG faces two significant challenges: information dilution and hallucination \cite{gao2023retrieval}. Information dilution arises when the volume of data is so large that it compromises the specificity and accuracy of the responses. Hallucination, on the other hand, refers to instances when the model produces outputs that are linguistically coherent but factually inaccurate or irrelevant to the input query \cite{bender2021dangers}.

    RAG can effectively handle domain-specific question answering using vectorization. Initially, the dataset is stored in a vector database using an embedding model. Embedding models transform text into numerical representations, or vectors, by capturing semantic meanings and relationships between words, which will allow for efficient similarity search later. When a user submits a query, it also converts into a query vector using the same embedding model. This ensures that both the dataset and the query are represented in comparable vector space. The retrieval component then matches the query vector against the database vector using similarity measures such as cosine similarity. The actual text associated  with the closest vectors is retrieved using the retrieval component of RAG. Then LLm uses this retrieved text as context to generate a response.

    This paper makes two key contributions: a) preparing domain-specific dataset to enhance the performance of RAG system b) The development of a novel modified RAG system, named QuIM-RAG (Question-to-question Inverted Index Matching), which introduces an inverted question matching approach with a quantized embedding index to improve the precision and efficiency of the information retrieval and answer generation processes. The domain-specific dataset is used to enhance the performance of RAG with a focus on data quality, relevance, and verifiable sources. We develope a comprehensive methodological approach for the data preparation phase, which included systematic data collection, cleaning, and structured organization. This approach is designed to improve the quality and reliability of the dataset corpus. The goal is to convert unprocessed data collected from websites into manageable chunks, which are then transformed into potential questions with the help of gpt-3.5-turbo-instruct model. A proper custom prompt is used to ensure that all the information available in each chunk is covered in custom corpus. To implement our modified RAG system,  we construct an inverted index based on the embedding of generated questions. Each question is transformed into an embedding vector, which is then quantized to the nearest prototype to reduce the computational load and facilitate efficient retrieval. The quantization process is key to building the inverted index, which efficiently links each prototype to related text chunks for fast retrieval. We deploy our system on the open-source Meta-LLaMA3-8B-instruct model. Throughout this implementation, our focus has been on maintaining the integrity and trustworthiness of the responses achieved through our system. 

    When a user submits a query, the query is initially converted into a vector using the same encoding technique employed for the previously generated questions. This vector is then quantized to the closest prototype in the embedding space, determined through cosine similarity. An inverted index is utilized to efficiently retrieve all embedding vectors associated with this prototype. For each of these vectors, the corresponding original question is decoded that reveals the associated text chunk from our dataset. Each chunk is not only rich in content but is also associated with a source link, providing direct access to the original documents. These chunks correspond to the sections of documents that are most relevant to the user's query. Offering direct access to the sources fulfills two key objectives:  a) it establishes a solid foundation for the trustworthiness of responses, and b) it encourages users to further explore and verify the information themselves to get a more interactive and reliable user experience. Instead of simply returning these chunks, our system uses them as context to generate a coherent response. This process leverages Meta-LLaMA3-8B-instruct model to produce an answer that is not only contextually aligned with the query but also substantiated by factual content from the original documents. It ensures the accuracy and relevance of the responses provided.     

    To evaluate the effectiveness of our dataset creation methodology and QuIM-RAG, we conduct a comprehensive assessment using the North Dakota State University (NDSU) website as a case study. The research question we are focusing on is:

    \textbf{Question 1:} "How does creating a web-retrieval dataset in a specific way impact the accuracy of responses generated by an LLM-powered system?"

    \textbf{Question 2:} "How does an advanced RAG model with a novel retrieval mechanism perform relative to a conventional RAG model?"

    In response, we perform a thorough comparison of the performance between our LLM-powered modified RAG system and a traditional RAG system, employing both a conventional web retrieval dataset and our custom dataset. The results from our comparative analysis highlight significant and considerable improvements in accuracy with the use of the custom dataset. Moreover, we aim to validate the effectiveness of our novel retrieval mechanism in enhancing the precision and reliability of the overall RAG system.

    \section{Related work}

    The development of Large Language Models (LLMs) has significantly transformed Natural Language Processing (NLP) in their ability to process and generate human-like text. State-of-the-art models like OpenAI's GPT series, Google's PaLM, and Meta's LLAMA series primarily utilize the Transformer architecture, which underpins their sophisticated understanding and generation of language \cite{radford2019language, chowdhery2023palm, touvron2023llama, vaswani2017attention}. These models showcase diverse architectural strategies, including exclusive use of decoders (as in GPT-2 and GPT-3), encoders (such as BERT and RoBERTa), or a blend of both in encoder-decoder frameworks (like BART), highlighting their versatility in approaching various linguistic tasks.

    Providing accurate answers to a given question using LLMs has become one of the most explored research areas in the last couple of years \cite{hadi2023survey, huo2023retrieving, Fan2023ABR, bolotova2022non, rogers2023qa}. The evaluation of Q\&A research has significantly shifted towards domain-specific applications, particularly in complex fields such as law, medicine, and education \cite{Ruder2021MultiDomainMQ}. While LLMs have made significant advances in generating human-like responses, they sometimes fall short in areas requiring specialized knowledge and are prone to producing inaccurate information, a phenomenon known as hallucination \cite{ji2023survey}. This limitation often arises because LLMs are typically trained on broad, general datasets that may not cover niche topics extensively. One way of overcoming this limitation is to use RAG for better performance in Q\&A \cite{mialon2023augmented, lazaridou2022internet}. Additionally, the introduction of Probably Asked Questions (PAQ) and the QA-pair retriever, RePAQ, highlights the evolving approaches to enhance LLMs' efficiency in Q\&A, aiming to bridge the gap in knowledge coverage through QA-pairs \cite{lewis2021paq}.

    In the context of RAG, it is crucial to efficiently retrieve relevant documents from the data source \cite{gao2023retrieval}. For that, data curation and preparation are important. RAG has been used for Open-Domain Question Answering (ODQA) using publicly available datasets that are easily accessible. However, few works have been done where custom datasets were required through web retrieval, content extraction, and segmentation \cite{sharma2023reliable, siragusa2023conditioning, kulkarni2024reinforcement, campese2023quadro}. A major challenge with RAG is its tendency to hallucinate with large text volumes. Our study suggests creating a structured Q\&A dataset from web text to enhance RAG's accuracy and reduce misinformation.

    \section{Methodology}

    Given a specific natural language question, the QA problem for a limited corpus is to identify and extract the most relevant and accurate answer from a predefined and constrained set of documents. The challenge lies in efficiently finding the most relevant answer that fits the question, taking into account the limited resources. Our key contribution is to pose finding the relevant answer as a matching process between potential questions that could have been asked for a document chunk with the actual user question. In this section, we first formalize the QA problem and then present our inverted question matching approach.
    \subsection{QA Problem for Limited Corpus}
    Given a corpus \( C \) consisting of a finite set of documents \(\{D_1, D_2, \ldots, D_n\}\) and a natural language question \( Q \), the task is to find the most relevant answer \( A \) from the corpus. Each document \( D_i \) in the corpus \( C \) contains a set of chunks (sentences) \(\{S_{i1}, S_{i2}, \ldots, S_{im}\}\). We only focus on natural language queries that seek specific information contained within the corpus. The output is a text snippet response derived from the corpus, that directly addresses the question \( Q \). \( A \) is a phrase generated from a combination of multiple pieces of information from the corpus. Hence the goal is to produce answer \( A \) such that \( A \) maximizes the relevance to \( Q \) within the constraints of \( C \).
    
    \[
    A = \arg\max_{A' \in C} \text{Relevance}(A', Q)
    \]
    
    Here relevance is a scoring function that quantifies how well \( A' \) answers \( Q \) using context from corpus \( C \). That might be computed using methods such as exactly matching expected keywords, using embeddings to measure the similarity between the question and potential answers, or ensuring that the answer fits logically with the surrounding text in the document. These natural solutions are either ineffective or prohibitive to implement. Our approach addresses this by generating hypothetical questions for each chunk \( S_i \) and finding the best match for the actual question to find the most likely chunk that answers the user query.
    
    \subsection{Inverted Question Matching to Find Relevant Chunks for Answer}
    Relevant chunks from various documents in the corpus can be combined to build the relevant informational context to answer the user’s question. Here we describe our inverted index scheme for matching document chunks in embedding space.
    
    \subsubsection*{Corpus and Document Structure}
    The corpus \( C \) consists of a collection of documents \( D_1, D_2, \ldots, D_n \). Each document \( D_i \) is composed of a set of chunks or sentences \( S_1, S_2, \ldots, S_m \). These chunks represent the fundamental units of information within each document, and they serve as the basis for generating semantically meaningful questions that capture the content of the text.
    
    \subsubsection*{Question Generation from Chunks}
    For each chunk \( S_j \) within a document \( D_i \), an instruction-following large language model (LLM) is employed to generate a set of questions \(\{q_{ij1}, q_{ij2}, \ldots, q_{ijk}\}\). These questions are designed to encapsulate the key information or concepts contained in the chunk \( S_j \). The process of question generation is crucial as it translates the raw text into a set of queries that can be later used for effective document retrieval. Intuitively, these can be viewed as "frequently asked questions" for the given chunk.
    
    \subsubsection*{Embedding of Generated Questions}
    Once the questions \(\{q_{ij1}, q_{ij2}, \ldots, q_{ijk}\}\) are generated, each question \( q_{ijl} \) is transformed into an embedding vector \( v_{ijl} \in \mathbb{R}^d \) using a pre-trained encoding scheme. These embedding vectors capture the semantic meaning of the questions in a high-dimensional space, enabling the system to compare and match questions based on their content.
    
    \subsubsection*{Quantization of Embeddings}
    Given the high dimensionality of the embedding space, direct comparison of all vectors would be computationally expensive. To mitigate this, each embedding vector \( v_{ijl} \) is quantized to the nearest prototype \( p_l \in \{p_1, p_2, \ldots, p_k\} \). The quantization is performed by finding the prototype \( p_l \) that minimizes the cosine similarity distance to \( v_{ijl} \):
    \[
    p_l = \arg\min_p \text{CosineSimilarity}(v_{ijl}, p)
    \]
    This quantization step reduces the complexity of the index and facilitates efficient matching of user queries.
    
    \subsubsection*{Construction of the Inverted Index}
    An inverted index \( I \) is built to map each prototype \( p_l \) to a set of embedding vectors \( v_{ijl} \) and their corresponding text chunks \( S_j \). The index is constructed as follows:
    \[
    I(p_l) = \{(v_{ijl}, S_j) \text{ for all } v_{ijl} \text{ quantized to } p_l\}
    \]
    This index enables fast lookup of relevant text chunks based on the quantized embeddings.
    
    \subsubsection*{Query Processing and Matching}
    When a user submits a query \( Q \), it is first embedded into a vector \( V \) using the same encoding scheme applied to the generated questions. The vector \( V \) is then quantized to the nearest prototype \( p_l \) in the embedding space using cosine similarity:
    \[
    p_l = \arg\min_p \text{CosineSimilarity}(V, p)
    \]
    The inverted index \( I \) is then used to retrieve the set of embedding vectors \( v_{ijl} \) that were quantized to \( p_l \).
    
    \subsubsection*{Retrieval of Relevant Text Chunks}
    For each retrieved embedding vector \( v_{ijl} \), the corresponding original question \( q_{ijl} \) is decoded, and the associated text chunk \( S_j \) is identified. These text chunks \( S_j \) represent the parts of the documents that are most relevant to the user query \( Q \).

\begin{algorithm}
\caption{Inverted Index in Embedding Space with Question Generation and Matching}
\begin{algorithmic}[1]
\State \textbf{Input:} Corpus \( C = \{D_1, D_2, \ldots, D_n\} \), Prototypes \( \{p_1, p_2, \ldots, p_k\} \)
\State \textbf{Output:} Relevant text chunks for user query \( Q \)

\Function{GenerateQuestions}{$S_i$}
    \State Use an Instruction-following LLM to generate questions \( \{q_{i1}, q_{i2}, \ldots, q_{ik}\} \) from chunk \( S_i \)
    \State \Return \( \{q_{i1}, q_{i2}, \ldots, q_{ik}\} \)
\EndFunction

\Function{EmbedQuestions}{$\{q_{i1}, \ldots, q_{ik}\}$}
    \State Initialize an empty list of embeddings
    \For{each question \( q_{ij} \) in \( \{q_{i1}, \ldots, q_{ik}\} \)}
        \State Compute embedding \( v_{ij} \leftarrow \text{Encode}(q_{ij}) \)
        \State Append \( v_{ij} \) to the list of embeddings
    \EndFor
    \State \Return \( \{v_{i1}, v_{i2}, \ldots, v_{ik}\} \)
\EndFunction

\Function{QuantizeEmbedding}{$v_{ij}, \{p_1, \ldots, p_k\}$}
    \State \Return Prototype \( p_l \leftarrow \arg\min_{p} \text{CosineSimilarity}(v_{ij}, p) \)
\EndFunction

\Function{BuildInvertedIndex}{$C, \{p_1, \ldots, p_k\}$}
    \State Initialize an empty inverted index \( I \)
    \For{each document \( D_i \) in \( C \)}
        \For{each chunk \( S_j \) in \( D_i \)}
            \State \( \text{Questions} \leftarrow \textbf{GenerateQuestions}(S_j) \)
            \State \( \text{Embeddings} \leftarrow \textbf{EmbedQuestions}(\text{Questions}) \)
            \For{each embedding \( v_{ij} \) in Embeddings}
                \State \( p_l \leftarrow \textbf{QuantizeEmbedding}(v_{ij}, \{p_1, \ldots, p_k\}) \)
                \State Update \( I(p_l) \leftarrow I(p_l) \cup \{(v_{ij}, S_j)\} \)
            \EndFor
        \EndFor
    \EndFor
    \State \Return \( I \)
\EndFunction

\Function{MatchQuery}{$Q, \{p_1, \ldots, p_k\}, I$}
    \State Compute query embedding \( V \leftarrow \text{Encode}(Q) \)
    \State Find the nearest prototype \( p_l \leftarrow \arg\min_{p} \text{CosineSimilarity}(V, p) \)
    \State Retrieve relevant embeddings \( E \leftarrow I(p_l) \)
    \State Initialize an empty list of relevant texts
    \For{each \( (v_{ij}, S_j) \) in \( E \)}
        \State Decode \( v_{ij} \) to original question \( q_{ij} \)
        \State Append chunk \( S_j \) to the list of relevant texts
    \EndFor
    \State \Return Relevant text chunks
\EndFunction

\end{algorithmic}
\end{algorithm}

    \subsection{Answer Generation for User Query}
    the relevant text chunks are returned to the large language model (LLM) to generate answer. However, in our scheme we use those chunk as the ”context” for generating a coherent response that is directly based on the query and supported by factual information from the original document.

    \section{Experimental Setup}

    Our work started with detailed data collection directly from two primary sources within the North Dakota State University (NDSU) domain: the NDSU Career Advising (\url{https://career-advising.ndsu.edu}) and the NDSU Catalog (\url{https://catalog.ndsu.edu}) websites. These websites were chosen due to their comprehensive and versatile information. While we aimed to gather all available information from both sites, the Career Advising site predominantly offered insights into career guidance, job search strategies, assistance with resume tailoring, interview preparation, and details on club activities, including their missions, visions, and current engagements. Simultaneously, the Catalog website provided us with comprehensive information on academic aspects such as admissions, enrollments, detailed course descriptions, listings of required and elective courses for each academic program, graduation requirements, and academic rules and policies.

    We employed advanced web scraping techniques, beginning with the main pages of NDSU-Career Advising and NDSU-Catalog. We then systematically navigated through their related sub-pages for comprehensive coverage of all available content on these sites. This methodology is structured into three main phases: Data Preparation, Retrieval, and Generation.

    \subsection{Data Preparation}

    \begin{figure}[h]
        \centerline{\includegraphics[width=0.7\textwidth]{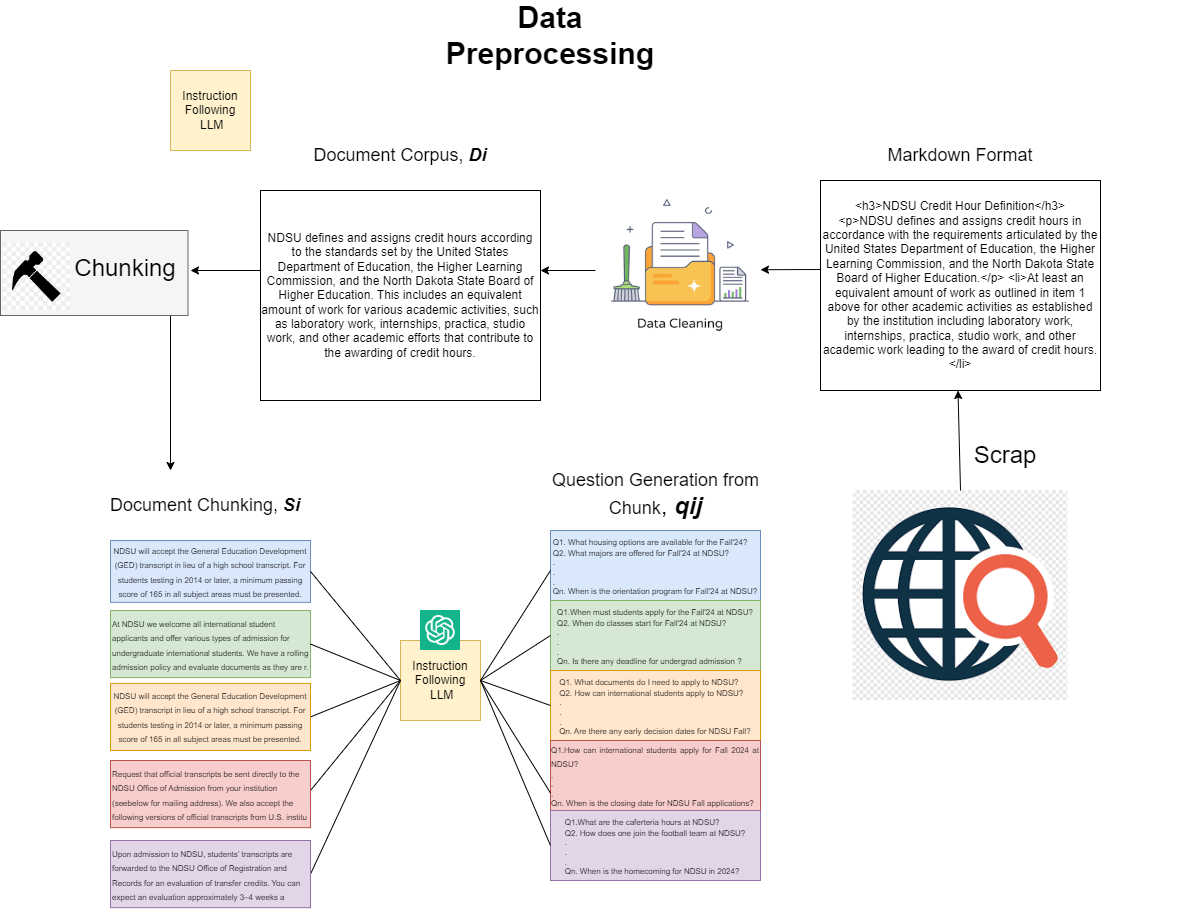}}
        \caption{Overall Architecture of Corpus Preparation for Modified RAG}
        \label{fig:data preparation}
    \end{figure}

    To create an accurate and complete dataset [Figure \ref{fig:data preparation}], we have constructed a customized web crawler integrating the BeautifulSoup and Scrapy frameworks. These two frameworks helped to traverse two primary sub-domains: NDSU Career Advising and NDSU Catalog. Starting from these parent links, the scraping process was designed to first identify and collect all associated page links reachable from these initial entry points. After collecting links, all HTML source code was pulled with the help of the crawler. Some data post-processing was done by removing the website header, footer, and unnecessary HTML tags to produce a cleaner version of the data. We also filtered out pages with less than 250 characters and those with a "404 page not found" title to ensure a relevant and high-quality dataset. The collection of all accessible links and their content offered a full insight into the website's resources and helped develop a broader dataset.

    The next phase involved deploying an instruction following large language model (LLM) to construct a targeted data corpus, for which we utilized GPT 3.5-turbo-instruct provided by OpenAI. This model is designed to interpret and execute instructions seamlessly. We employed TikToken for chunking the data into 1000 tokens and creating overlapping chunks of 200 characters to ensure that the model comprehended the context of each chunk effectively. For each distinct chunk, a set of questions is generated to encapsulate the key information of the chunk. These questions are later used for effective context retrieval and help with accurate answer generation. The number of question-answer pairs generated for each context varies. This methodical approach enabled us to develop a customized dataset for our RAG system to ensure thorough coverage of each topic.
    
     During the development of our data corpus, we placed great emphasis on ensuring that each set of questions is both accurate and relevant to the chunk. To achieve this, we implemented a detailed manual review process following the initial automated generation of questions. This review was conducted by two researchers who closely examined each set to confirm its factual accuracy and contextual relevance. During this review, we identified instances where the automated system generated questions that did not adequately address the chunk or were incomplete. We reprocessed these sections to resolve any inconsistencies and ensure that each set accurately mirrored the data it was derived from. We also eliminated duplicates and irrelevant questions to maintain the integrity and value of the corpus. Furthermore, we carefully check for semantic and syntactic errors to ensure information correctness and grammatical accuracy. This enhanced the overall reliability of how we prepare the dataset.

    \begin{table}[ht]
        \centering
        \setlength{\tabcolsep}{12pt}
        \begin{tabular}{|l|c|c|}
            \hline
            \textbf{Metric} & \textbf{NDSU Catalog} & \textbf{NDSU Career} \\ 
              & \textbf{Dataset} & \textbf{Advising}    \\ \hline
            No. of Links Scraped      & 288   & 399    \\ \hline
            No. of Questions      & 9,027   & 13,582    \\ \hline
            No. of chunks      & 296   & 435    \\ \hline
        \end{tabular}
        \caption{Summary of the NDSU Catalog and Career Advising Datasets}
        \label{table:ndsucareerdataset}
    \end{table}

    Table \ref{table:ndsucareerdataset} provides an overview of the dataset used in the paper. The data was obtained by scraping 687 links associated with North Dakota State University's (NDSU) career-advising and catalog resources. The scraped context was then formed into manageable chunks with set of questions and links using GPT-3.5-turbo-instruct model. This resulted in 9,027 questions from the NDSU Catalog dataset and 13,582 questions from the NDSU Career Advising dataset, resulting in a total of 22,609 questions.

    \subsection{Ground Truth Preparation}
    We prepared ground truth data to ensure high accuracy and relevance for evaluating our RAG system. The process began by selecting relevant content from the university website, which served as potential sources of information. We extracted this content using automated web scraping tools to preserve all essential details. we then manually curated this content to verify its factual accuracy and direct relevance to potential user queries. Based on this verified content, we created question-answer pairs that accurately reflect the information available on the website. We ensured that each pair was complete and directly linked to its original source. This prepared ground truth helps to effectively assess the performance of our RAG system.

    \subsection{Inverted Index Contruction for Question matching}

    \begin{figure}[htbp]
        \centerline{\includegraphics[width=0.7\textwidth]{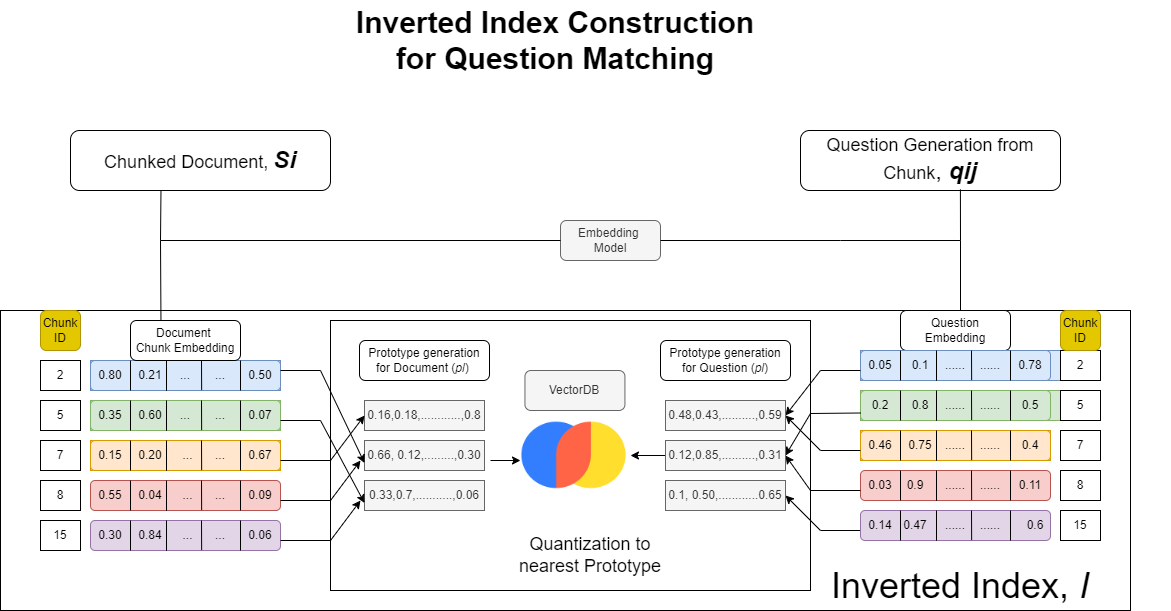}}
        \caption{Illustration of Inverted Index Construction for Question Matching}
        \label{fig:ingestion}
    \end{figure}

    After constructing the dataset corpus, the next step involves preparing an inverted index to facilitate efficient question matching (Figure \ref{fig:ingestion}). Once the questions are generated, we use an embedding vector to capture the semantic meaning of each chunk and data corpus. We employed a pretrained embedding model from HuggingFace (model name: BAAI/bge-large-en-v1.5) for its high performance on the Multi-Task Evaluation Benchmark (MTEB) \cite{muennighoff2022mteb}. This model utilizes a flag embedding technique, which can excel in the semantic search and retrieval capability of any LLM. The vector representations for both documents and questions are subsequently quantized into the nearest prototype vectors in a high-dimensional space. We use chromaDb as a vector database for the quantization process that stores prototype vectors for both document chunks and generated questions. The vectors are represented in a format where each prototype vector is derived by minimizing the cosine similarity distance between the chunk or question embedding and available prototypes to effectively reduce the computational complexity of direct vector comparisons. These quantized embeddings are then formed in an inverted index, which maps each prototype vector to the corresponding text chunks and their derived questions. This inverted index serves as the backbone of the retrieval process. To efficiently retrieve questions and corresponding chunks, we use chunk ids in chunk vectorDB and question vectorDB.

    \subsection{Retrieval}

    \begin{figure}[htbp]
        \centerline{\includegraphics[width=0.7\textwidth]{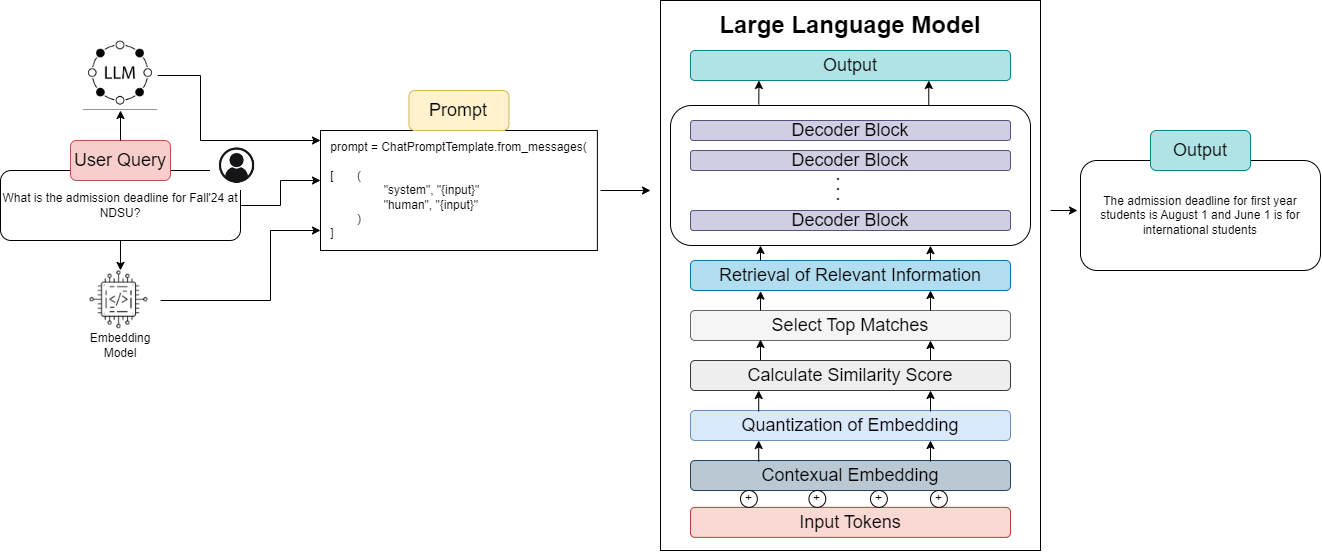}}
        \caption{Overall Retrieval and Generation Architecture for RAG}
        \label{fig:retrieval}
    \end{figure}

    The retrieval system is designed to efficiently match user queries with the most relevant information within a comprehensive knowledge base. When a user submits a query \(Q\), our system initiates by transforming \(Q\) into an embedding vector using the same embedding process that applied to data corpus. Then, from the inverted index, it matches for the closest semantic question from data corpus by similarity search using cosine similarity. It selects the top 3 questions (\(k=3\)) that best align semantically with the query. Then the system retrieved the associated chunks related to these questions and decode and return these as context to the llAma3-7b-instruct.

    \subsection{Generation}
    The final response to the user is generated by an open-source large language model (LLM) from Hugging Face named Llama3-8b-instruct. This is the latest addition to the Llama series while we were writing the paper, with significant advancements in AI capability. We have used the instruct-tuned version for its powerful conversational and chat capability. Llama3 supports a context length of 8,000 tokens, which allows for more extended interactions and more complex input handling compared to many previous models. Longer context capacity ensures that the model can consider a wider range of information to deliver the most relevant and coherent responses.

    In the generation process, the model retrieves the most relevant context from the retrieval model. Then it integrates the information seamlessly using a custom prompt (Figure \ref{prompt}). Since the Llama3 model is trained on a vast database from diverse sources, it has a broad understanding of generating human-like text. This training enables the model to effectively utilize the retrieved information \(\{a_i\}\) to generate a response that is not only relevant but also contextually coherent. The final response is then delivered back to the user as output.

    \subsection{Prompt}

    \begin{figure}[tp]
        \centering
        \includegraphics[width=0.7\textwidth]{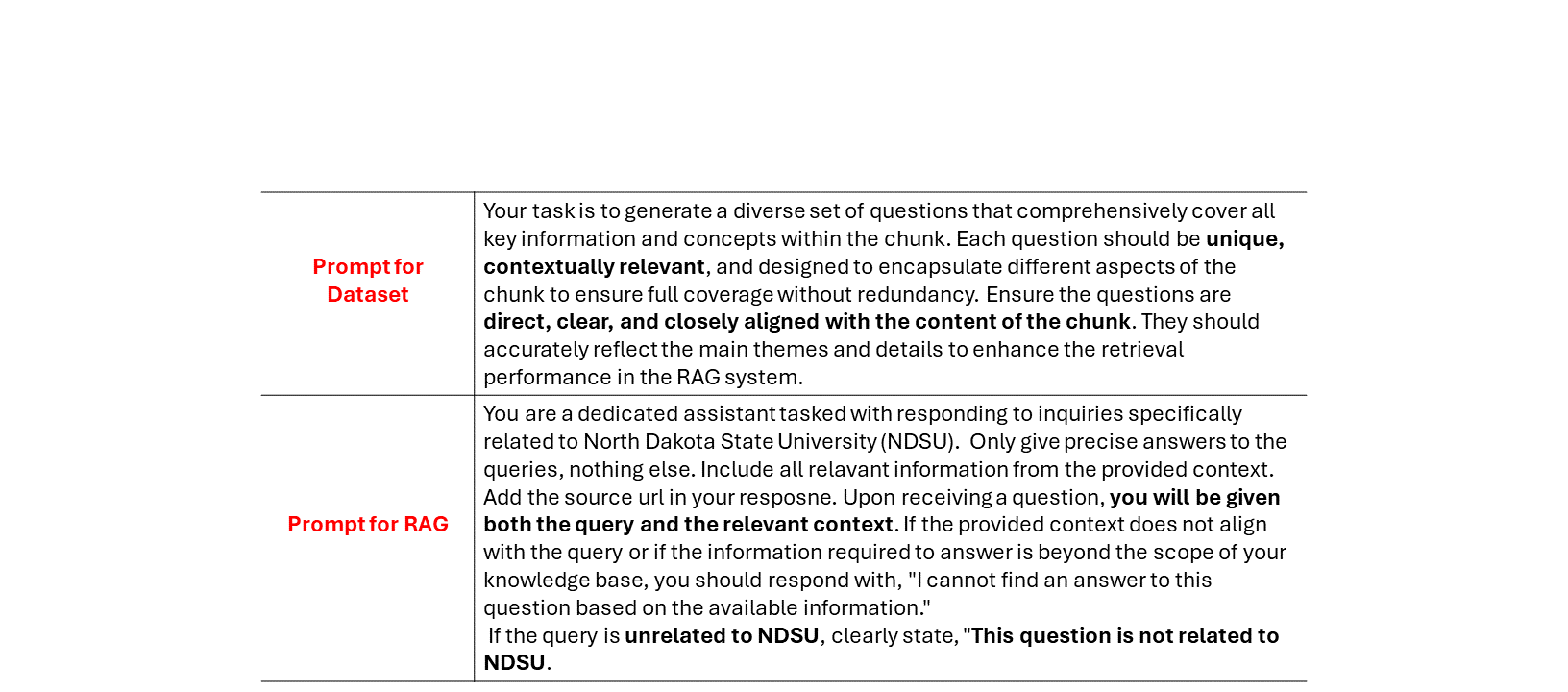}
        \caption{The upper section details a prompt designed for creating a custom dataset, focusing on generating set of questions for each chunk. The lower section outlines a prompt for a RAG system, emphasizing accuracy and directive responses based on the dataset, with instructions on how to handle queries that extend beyond available data.}
        \label{prompt}
    \end{figure}

    To refine the capabilities of RAG, developing a comprehensive dataset is essential. This dataset is crucial for training the RAG system to deliver precise and contextually relevant responses. Acknowledging the necessity to encompass the breadth of information within our dataset, we formulated a custom prompting strategy. This critical step ensures that the resultant dataset of question-and-answer (Q\&A) pairs is not only comprehensive but also precise and reflective of the diverse content within the source texts.

    Our goal is to achieve comprehensive coverage of the chunk and ensure the diversity and accuracy of generated set of questions. The designed prompt [figure \ref{prompt}] explicitly instructs the model to generate a set of questions that cover all the key information of each chunk. The prompt directs the model to avoid redundancy, ensuring that each question is unique and contextually relevant to the text. This approach helps prevent the generation of overlapping questions and ensures that each query contributes distinct value to understanding the chunk's content.

    Building upon this foundation, it is imperative to address how our RAG model manages queries that extend beyond the confines of its knowledge base. To tackle different types of questions effectively and prevent the provision of inaccurate responses to out-of-domain (OOD) inquiries, another prompting strategy was implemented for the RAG model itself. In instances where a query seeks information absent from the text segments known to the model, the model is prompted to clearly state its inability to provide a relevant answer. This ensures the model explicitly knows its boundaries and helps prevent the provision of false or invented responses.

    \subsection{Comparison to Traditional RAG System}
    Traditional RAG systems typically follow a straightforward "Retrieve-Read" methodology \cite{gao2023retrieval}. These systems begin with an indexing phase where data is segmented into vectorized chunks. These chunks are then retrieved in response to user queries based on their semantic similarity. However, these systems often struggle with information overload and accuracy, leading to responses that may not always be contextually appropriate or factually correct. This affects the overall quality of the responses they produce.
    In contrast, our advanced RAG system introduces substantial improvements in both the retrieval and generation phases to address these limitations. Our retrieval system is designed to match user queries with the most relevant information within a comprehensively curated knowledge base. By employing an enhanced semantic chunking method, our system segments data into coherent units that encapsulate complete thematic entities. The retrieval process in our advanced RAG system utilizes state-of-the-art embedding models that improve the traditional vectorization methods. It features a refined retrieval system that uses state-of-the-art embedding models for better semantic matching and segments data into coherent units to make the retrieved information highly relevant to user queries. The generation phase is powered by the Llama3-8b-instruct model from Hugging Face, which supports complex inputs and provides coherent outputs over extended conversational turns. Additionally, we've implemented a custom prompting strategy to ensure the generated Q\&A pairs are detailed and accurate and significantly minimizing issues like redundancy and hallucination commonly found in traditional RAG systems.

    \section{Evaluation}

      To evaluate the performance of our QuIM-RAG model, we employ two evaluation frameworks, BERTScore \cite{zhang2019bertscore} and RAGAS \cite{es2023ragas}. These frameworks will provide a detailed quantitative analysis that focuses on the semantic accuracy and relevance of the models response. We use BERTScore and RAGAS instead of more commonly used metrics such as BLEU \cite{s_2024}, METEOR \cite{banerjee2005meteor}, and ROUGE \cite{barbella2022rouge} which don't quite match our evaluation needs. BLEU and METEOR are traditionally applied to machine translation tasks, and ROUGE is tailored for text summarizing assessments. Given that our model operates within a RAG-based question-answering context, these conventional metrics fall short in accurately measuring its performance. This limitation has led us to select evaluation methods that are more appropriate and tailored to our model's specific needs.

    \subsection{BERTScore}

    BERTScore is used to evaluate the semantic quality of text by measuring the overlap between model-generated outputs and reference texts \cite{zhang2019bertscore}. Unlike traditional metrics that rely on n-gram overlaps, BERTScore utilizes contextual embedding to capture deeper semantic meanings to it.  It is highly effective for tasks where linguistic precision is crucial. BERTScore operates by first transforming both the reference, $\mathbf{x} = (x_1, x_2, \dots, x_n)$, and candidate, $\hat{\mathbf{x}} = (\hat{x}_1, \hat{x}_2, \dots, \hat{x}_m)$ texts into contextual embedding using models like BERT or RoBERTa[]. 
    
    \[
    \text{BERT}((x_1, x_2, \dots, x_n)) = (X_1, X_2, \dots, X_n)
    \]
    \[
    \text{BERT}((\hat{x}_1, \hat{x}_2, \dots, \hat{x}_m)) = (\hat{X}_1, \hat{X}_2, \dots, \hat{X}_m)
    \]
    
    These embeddings integrate the surrounding context of words to provide a robust representation of textual meaning. The core of BERTScore's evaluation method is the pairwise cosine similarity calculation between tokens of the reference text and the candidate text. For each token in the reference, it identifies the token in the candidate text that has the highest cosine similarity and vice versa.
    
    \[
    \text{similarity}(X_i, \hat{X}_j) = \frac{X_i^T \hat{X}_j}{\|X_i\| \|\hat{X}_j\|}
    \]
    
    To evaluate the quality of text generation, three distinct metrics are used: BERT-Precision, BERT-Recall, and BERT-F1.
    
    BERT-Precision(BERT-P): This metric quantifies the semantic similarity between each token in the generated text and the nearest equivalent token in the reference text. By employing greedy matching, BERT-P optimizes the similarity scores to effectively handle the linguistic variability of semantic words that are interchangeable without changing the overall meaning. Greedy matching is critical in the language domain because multiple words may hold similar meanings to the ground truth, and the words of sentences can be structured in various ways without changing identical semantic content.

    {\small
    \[
    P_{\text{BERT}} = \frac{1}{|\hat{\mathbf{x}}|} \sum_{\hat{x}_j \in \hat{\mathbf{x}}} \max_{x_i \in \mathbf{x}} \left\{ \text{cosine similarity} (\hat{X}_i^T, X_j) \right\}_{\text{\footnotesize{greedy matching}}}
    \]
    }

    BERT-Recall (BERT-R): It measures the model's completeness by comparing each token of reference text with generated text that has highest cosine similarity. The recall score is computed by averaging these maximum similarity scores. 

    {\small
    \[
    R_{\text{BERT}} = \frac{1}{|\mathbf{x}|} \sum_{x_i \in \mathbf{x}} \max_{\hat{x}_j \in \hat{\mathbf{x}}} \left\{ \text{cosine similarity} (X_i^T, \hat{X}_j) \right\}_{\text{\footnotesize{greedy matching}}}
    \]
    }

    BERT-F1 Score (BERT F1): Combining  the BERT-P and Bert-R, the F1 score provides a holistic measure of text generation quality by balancing both the breadth and depth of semantic content captured in the generated text.
    \[
    F_{\text{BERT}} = 2 \times \frac{P_{\text{BERT}} \times R_{\text{BERT}}}{P_{\text{BERT}} + R_{\text{BERT}}}
    \]

    \begin{figure*}[tp]
        \centerline{\includegraphics[width=\textwidth]{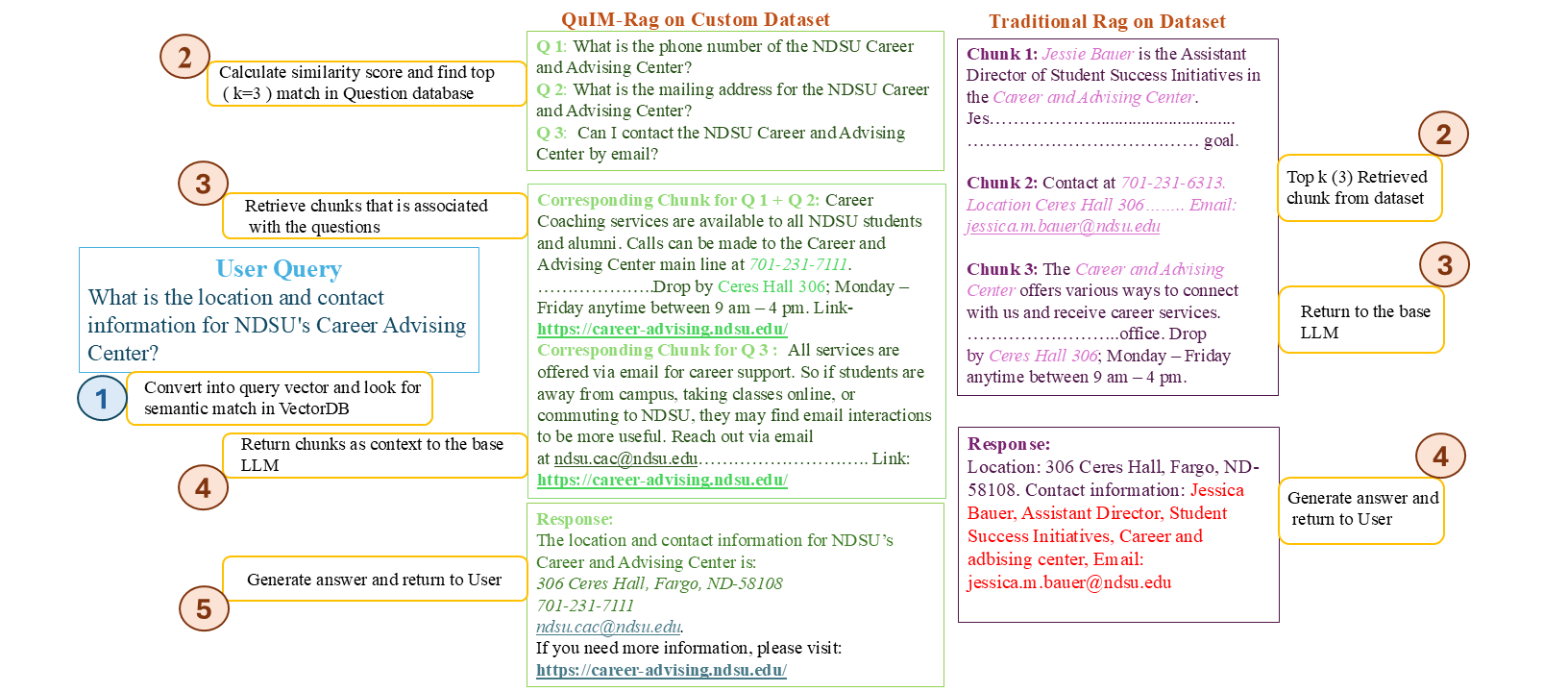}}
        \caption{Workflow of QuIM-RAG system and Traditional RAG system for User Query Processing}
        \label{fig:Chat_custom.png}
    \end{figure*}

\subsection{Retrieval-Augmented Generation Assessment (RAGAS)}

RAGAS \cite{es2023ragas} is designed to evaluate the effectiveness of RAG systems that combine retrieval mechanisms with LLMs to provide accurate information. It evaluates these systems by focusing on the precision of context retrieval, the reliability of the generated content, and the overall quality of the text output. Importantly, RAGAS is constructed in such a way that it eliminates the need for human annotations in the evaluation process. Since Ragas is integrated with popular frameworks like LangChain and LLama-index, developers can conveniently utilize and assess their systems using this framework.

\subsubsection{Evaluation Metrics}
RAGAS evaluates three primary aspects of RAG architectures:

\begin{itemize}
    \item \textbf{Faithfulness:} This metric evaluates how well the retrieval system grounds the generated responses in the provided context. It involves extracting statements from the generated answers and verifying each statement against the retrieved context using a validation function to determine if the context supports the statement.
    \[
    \text{Faithfulness} = \frac{\text{Number of verified claims}}{\text{Total claims made}}
    \]
    
    \item \textbf{Answer Relevance:} This metric evaluates the appropriateness of the generated answers to address the posed questions. It is measured by generating potential questions from the answers using an LLM and calculating the cosine similarity of these questions to the original question. The relevance score is the average of these similarities, which indicates the directness and appropriateness of the answers.

    \[
    AR = \frac{1}{n} \sum_{i=1}^n \text{sim}(q, q_i)
    \]
    where \( q_i \) are questions generated based on the answer and sim is the cosine similarity between embeddings of the original question \( q \) and \( q_i \).

    \item \textbf{Context Relevance:} This metric determines whether the retrieved context contains primarily relevant information needed to answer the question. The process involves extracting sentences from the context that are necessary to answer the question and then calculating the ratio of these extracted sentences to the total number of sentences in the context.

   \[
    CR = \frac{\text{Number of relevant sentences extracted}}{\text{Total sentences in the context}}
    \]
 \end{itemize}

    \section{RESULT}

    The efficiency of our RAG system are demonstrated in Figure \ref{fig:Chat_custom.png}, which offers a detailed comparison between responses generated from traditional and custom datasets when queried about the location and contact information of the North Dakota State University (NDSU) Career and Advising Center. As shown in figure \ref{fig:Chat_custom.png}, when a user submits a query, it is first converted into a query vector. The encoded query vector is then used to compute similarity scores against a database of question vectors, which represent pre-stored questions associated with various chunks of information. The system identifies the top k matches (in this case, k=3), which are the questions most semantically related to the user query based on their similarity scores. For each of the top matching question vectors, the system retrieves the associated text chunks from the dataset. In this scenario, question 1 and 2 comes from the same chunk and question 3 comes from another chunk. These chunks contain the detailed information corresponding to each question. The retrieved chunks serve as context for the base LLM Meta-LLaMA3-8B-instruct to process this context along with the user query to generate a coherent and detailed answer that addresses the user’s specific needs. This approach directly addressed the query with precise information, including the phone number, mailing address, and email contact for the career and advising center. Additionally, it provides the source link for users who wish to further explore or verify the information. The traditional dataset response is also begins with similar query processing steps. However, when querying for contact information, the traditional dataset yields the details of an Assistant Director, which, while correct, may not align with the general inquiry intentions of users seeking contact information. The custom dataset and QuIM-RAG help the LLM to understand users' specific needs and avoid providing excessively detailed or irrelevant information, which is a common issue with the traditional RAG system. The comparison highlights that the QuIM-RAG with custom dataset provides more targeted and relevant response to the user.

    \begin{table}[ht]
        \centering
        \begin{tabular}{|l|c|c|c|c|}
            \hline
            Evaluation Matrix & \multicolumn{2}{c|}{Traditional RAG} & \multicolumn{2}{c|}{QuIM-RAG} \\ \cline{2-5}
            & Traditional & Custom & Traditional & Custom \\ \hline % Abbreviated headings
            Faithfulness      & 0.69        & 0.72   & 0.91        & 1.00   \\ \hline
            Answer Relevancy  & 0.79        & 0.82   & 0.93        & 0.99   \\ \hline
            Context Precision & 0.45        & 0.69   & 0.82        & 0.92   \\ \hline
            Context Recall    & 0.39        & 0.45   & 0.60        & 0.74   \\ \hline
            Harmfulness       & 0           & 0      & 0           & 0      \\ \hline
            BERTScore P.      & 0.32        & 0.37   & 0.55        & 0.63   \\ \hline
            BERTScore R.      & 0.29        & 0.35   & 0.63        & 0.71   \\ \hline
            F1 Score          & 0.31        & 0.36   & 0.59        & 0.0.67 \\ \hline
        \end{tabular}
        \caption{Comparison of Performance Metrics for Traditional RAG and our novel QuIM-RAG using traditional and Custom Datasets}
        \label{table:rag_comparison}
    \end{table}

    The efficacy of the RAG system was evaluated across a spectrum of metrics to ascertain the impact of employing traditional versus custom datasets on performance. Table \ref{table:rag_comparison} presents a comparative analysis of these metrics for both Traditional RAG and our novel QuIM-RAG systems. For the traditional RAG using traditional datasets, faithfulness scores reflected a moderate accuracy of 0.69, while the implementation with custom datasets showed an improved score of 0.72. The QuIM-RAG system, however, demonstrated superior faithfulness, particularly with custom datasets, achieving a perfect score of 1.00. This trend was mirrored across other metrics, including Answer Relevancy and Context Precision, with the QuIM-RAG system outperforming the Traditional RAG, underscoring the benefits of custom data in enhancing system accuracy. The metrics of Context Recall and BERTScore precision and recall also exhibited significant improvements when transitioning from traditional to custom datasets within both RAG systems. Notably, the Traditional RAG system experienced a Context Recall increase from 0.39 to 0.45 with custom datasets, whereas our QuIM-RAG's recall improved from 0.60 to 0.74, suggesting a better ability to retrieve relevant information when utilizing custom data. A consistent score of zero for Harmfulness across all models indicates a successful avoidance of generating harmful content. This is a critical aspect, highlighting the systems' reliability and the effectiveness of underlying filtering mechanisms.

    BERTScores further support the findings, with QuIM-RAG system achieving a precision of 0.63 and a recall of 0.71 using custom datasets, in comparison to 0.32 and 0.29, respectively, with the traditional RAG on traditional data. The F1 Score, a harmonic mean of precision and recall, showcases a conclusive elevation from 0.31 in the traditional RAG with traditional data to 0.67 in the QuIM-RAG with custom data. This quantifiable advancement validates the integration of custom datasets as a substantial enhancement to the RAG system's performance.

    \section{Conclusion}

    This paper addressed challenges of using large language models (LLM) for domain-specific question-answering. By integrating an advanced retrieval-augmented generation (RAG) system and a methodological approach to data preparation has enhance the quality of responses generated by these systems. Creating a custom dataset specifically designed for the domain in question has been key in reducing common problems like information dilution and hallucination that often seen in traditional RAG systems when they handle large amounts of unstructured data. Our evaluations show that our novel QuIM-RAG system, leveraging a custom dataset and Llama-3-8b-instruct, improves the accuracy and relevance of its responses. It performs much better than the baseline dataset, which is made from raw web data. Additionally, the responses include source links with the user's query offers further opportunities for user to seek more information and verify the details provided. Given that university websites frequently update their content every semester, we are planning to design a content retrieval mechanism that updates its corpora every four months. This will ensure that any newly added content is incorporated into the system to keep the information up-to-date and relevant for users. Looking ahead, our future goal is to conduct a comprehensive user study to assess user satisfaction and system usability. The insights gained from this study will help us refine the system further to ensure that it continues to meet user needs effectively while remaining adaptable to evolving content.

    \section*{Acknowledgment}
        This work utilized resources from the Center for Computationally Assisted Science and Technology (CCAST) at North Dakota State University, supported by NSF MRI Award No. 2019077, including computational infrastructure and support services. These resources, including computational infrastructure and essential support services, were invaluable for the processing and analysis tasks involved in our research. We sincerely thank CCAST for their assistance.

    \bibliographystyle{IEEEtran}
    \bibliography{references}

\begin{thebibliography}{10}
\providecommand{\url}[1]{#1}
\csname url@samestyle\endcsname
\providecommand{\newblock}{\relax}
\providecommand{\bibinfo}[2]{#2}
\providecommand{\BIBentrySTDinterwordspacing}{\spaceskip=0pt\relax}
\providecommand{\BIBentryALTinterwordstretchfactor}{4}
\providecommand{\BIBentryALTinterwordspacing}{\spaceskip=\fontdimen2\font plus
\BIBentryALTinterwordstretchfactor\fontdimen3\font minus \fontdimen4\font\relax}
\providecommand{\BIBforeignlanguage}[2]{{%
\expandafter\ifx\csname l@#1\endcsname\relax
\typeout{** WARNING: IEEEtran.bst: No hyphenation pattern has been}%
\typeout{** loaded for the language `#1'. Using the pattern for}%
\typeout{** the default language instead.}%
\else
\language=\csname l@#1\endcsname
\fi
#2}}
\providecommand{\BIBdecl}{\relax}
\BIBdecl

\bibitem{brown2020language}
T.~Brown, B.~Mann, N.~Ryder, M.~Subbiah, J.~D. Kaplan, P.~Dhariwal, A.~Neelakantan, P.~Shyam, G.~Sastry, A.~Askell \emph{et~al.}, ``Language models are few-shot learners,'' \emph{Advances in neural information processing systems}, vol.~33, pp. 1877--1901, 2020.

\bibitem{wang2023empower}
Z.~Wang, F.~Yang, P.~Zhao, L.~Wang, J.~Zhang, M.~Garg, Q.~Lin, and D.~Zhang, ``Empower large language model to perform better on industrial domain-specific question answering,'' \emph{arXiv preprint arXiv:2305.11541}, 2023.

\bibitem{kandpal2023large}
N.~Kandpal, H.~Deng, A.~Roberts, E.~Wallace, and C.~Raffel, ``Large language models struggle to learn long-tail knowledge,'' in \emph{International Conference on Machine Learning}.\hskip 1em plus 0.5em minus 0.4em\relax PMLR, 2023, pp. 15\,696--15\,707.

\bibitem{ovadia2023fine}
O.~Ovadia, M.~Brief, M.~Mishaeli, and O.~Elisha, ``Fine-tuning or retrieval? comparing knowledge injection in llms,'' \emph{arXiv preprint arXiv:2312.05934}, 2023.

\bibitem{kirkpatrick2017overcoming}
J.~Kirkpatrick, R.~Pascanu, N.~Rabinowitz, J.~Veness, G.~Desjardins, A.~A. Rusu, K.~Milan, J.~Quan, T.~Ramalho, A.~Grabska-Barwinska \emph{et~al.}, ``Overcoming catastrophic forgetting in neural networks,'' \emph{Proceedings of the national academy of sciences}, vol. 114, no.~13, pp. 3521--3526, 2017.

\bibitem{goodfellow2013empirical}
I.~J. Goodfellow, M.~Mirza, D.~Xiao, A.~Courville, and Y.~Bengio, ``An empirical investigation of catastrophic forgetting in gradient-based neural networks,'' \emph{arXiv preprint arXiv:1312.6211}, 2013.

\bibitem{chen2020recall}
S.~Chen, Y.~Hou, Y.~Cui, W.~Che, T.~Liu, and X.~Yu, ``Recall and learn: Fine-tuning deep pretrained language models with less forgetting,'' \emph{arXiv preprint arXiv:2004.12651}, 2020.

\bibitem{Saha2024}
B.~Saha and U.~Saha, ``Enhancing international graduate student experience through ai-driven support systems: A llm and rag-based approach,'' in \emph{2024 International Conference on Data Science and Its Applications (ICoDSA)}, North Dakota State University.\hskip 1em plus 0.5em minus 0.4em\relax IEEE, 2024, pp. 300--304, in press.

\bibitem{lewis2020retrieval}
P.~Lewis, E.~Perez, A.~Piktus, F.~Petroni, V.~Karpukhin, N.~Goyal, H.~K{\"u}ttler, M.~Lewis, W.-t. Yih, T.~Rockt{\"a}schel \emph{et~al.}, ``Retrieval-augmented generation for knowledge-intensive nlp tasks,'' \emph{Advances in Neural Information Processing Systems}, vol.~33, pp. 9459--9474, 2020.

\bibitem{neelakantan2022text}
A.~Neelakantan, T.~Xu, R.~Puri, A.~Radford, J.~M. Han, J.~Tworek, Q.~Yuan, N.~Tezak, J.~W. Kim, C.~Hallacy \emph{et~al.}, ``Text and code embeddings by contrastive pre-training,'' \emph{arXiv preprint arXiv:2201.10005}, 2022.

\bibitem{gao2023retrieval}
Y.~Gao, Y.~Xiong, X.~Gao, K.~Jia, J.~Pan, Y.~Bi, Y.~Dai, J.~Sun, and H.~Wang, ``Retrieval-augmented generation for large language models: A survey,'' \emph{arXiv preprint arXiv:2312.10997}, 2023.

\bibitem{bender2021dangers}
E.~M. Bender, T.~Gebru, A.~McMillan-Major, and S.~Shmitchell, ``On the dangers of stochastic parrots: Can language models be too big?'' in \emph{Proceedings of the 2021 ACM conference on fairness, accountability, and transparency}, 2021, pp. 610--623.

\bibitem{radford2019language}
A.~Radford, J.~Wu, R.~Child, D.~Luan, D.~Amodei, I.~Sutskever \emph{et~al.}, ``Language models are unsupervised multitask learners,'' \emph{OpenAI blog}, vol.~1, no.~8, p.~9, 2019.

\bibitem{chowdhery2023palm}
A.~Chowdhery, S.~Narang, J.~Devlin, M.~Bosma, G.~Mishra, A.~Roberts, P.~Barham, H.~W. Chung, C.~Sutton, S.~Gehrmann \emph{et~al.}, ``Palm: Scaling language modeling with pathways,'' \emph{Journal of Machine Learning Research}, vol.~24, no. 240, pp. 1--113, 2023.

\bibitem{touvron2023llama}
H.~Touvron, L.~Martin, K.~Stone, P.~Albert, A.~Almahairi, Y.~Babaei, N.~Bashlykov, S.~Batra, P.~Bhargava, S.~Bhosale \emph{et~al.}, ``Llama 2: Open foundation and fine-tuned chat models,'' \emph{arXiv preprint arXiv:2307.09288}, 2023.

\bibitem{vaswani2017attention}
A.~Vaswani, N.~Shazeer, N.~Parmar, J.~Uszkoreit, L.~Jones, A.~N. Gomez, {\L}.~Kaiser, and I.~Polosukhin, ``Attention is all you need,'' \emph{Advances in neural information processing systems}, vol.~30, 2017.

\bibitem{hadi2023survey}
M.~U. Hadi, R.~Qureshi, A.~Shah, M.~Irfan, A.~Zafar, M.~B. Shaikh, N.~Akhtar, J.~Wu, S.~Mirjalili \emph{et~al.}, ``A survey on large language models: Applications, challenges, limitations, and practical usage,'' \emph{Authorea Preprints}, 2023.

\bibitem{huo2023retrieving}
S.~Huo, N.~Arabzadeh, and C.~L. Clarke, ``Retrieving supporting evidence for llms generated answers,'' \emph{arXiv preprint arXiv:2306.13781}, 2023.

\bibitem{Fan2023ABR}
\BIBentryALTinterwordspacing
L.~Fan, L.~Li, Z.~Ma, S.~Lee, H.~Yu, and L.~Hemphill, ``A bibliometric review of large language models research from 2017 to 2023,'' \emph{ArXiv}, vol. abs/2304.02020, 2023. [Online]. Available: \url{https://api.semanticscholar.org/CorpusID:257952516}
\BIBentrySTDinterwordspacing

\bibitem{bolotova2022non}
V.~Bolotova, V.~Blinov, F.~Scholer, W.~B. Croft, and M.~Sanderson, ``A non-factoid question-answering taxonomy,'' in \emph{Proceedings of the 45th International ACM SIGIR Conference on Research and Development in Information Retrieval}, 2022, pp. 1196--1207.

\bibitem{rogers2023qa}
A.~Rogers, M.~Gardner, and I.~Augenstein, ``Qa dataset explosion: A taxonomy of nlp resources for question answering and reading comprehension,'' \emph{ACM Computing Surveys}, vol.~55, no.~10, pp. 1--45, 2023.

\bibitem{Ruder2021MultiDomainMQ}
\BIBentryALTinterwordspacing
S.~Ruder and A.~Sil, ``Multi-domain multilingual question answering,'' \emph{Proceedings of the 2021 Conference on Empirical Methods in Natural Language Processing: Tutorial Abstracts}, 2021. [Online]. Available: \url{https://api.semanticscholar.org/CorpusID:245289877}
\BIBentrySTDinterwordspacing

\bibitem{ji2023survey}
Z.~Ji, N.~Lee, R.~Frieske, T.~Yu, D.~Su, Y.~Xu, E.~Ishii, Y.~J. Bang, A.~Madotto, and P.~Fung, ``Survey of hallucination in natural language generation,'' \emph{ACM Computing Surveys}, vol.~55, no.~12, pp. 1--38, 2023.

\bibitem{mialon2023augmented}
G.~Mialon, R.~Dess{\`\i}, M.~Lomeli, C.~Nalmpantis, R.~Pasunuru, R.~Raileanu, B.~Rozi{\`e}re, T.~Schick, J.~Dwivedi-Yu, A.~Celikyilmaz \emph{et~al.}, ``Augmented language models: a survey,'' \emph{arXiv preprint arXiv:2302.07842}, 2023.

\bibitem{lazaridou2022internet}
A.~Lazaridou, E.~Gribovskaya, W.~Stokowiec, and N.~Grigorev, ``Internet-augmented language models through few-shot prompting for open-domain question answering,'' \emph{arXiv preprint arXiv:2203.05115}, 2022.

\bibitem{lewis2021paq}
P.~Lewis, Y.~Wu, L.~Liu, P.~Minervini, H.~K{\"u}ttler, A.~Piktus, P.~Stenetorp, and S.~Riedel, ``Paq: 65 million probably-asked questions and what you can do with them,'' \emph{Transactions of the Association for Computational Linguistics}, vol.~9, pp. 1098--1115, 2021.

\bibitem{sharma2023reliable}
V.~Sharma and V.~Raman, ``A reliable knowledge processing framework for combustion science using foundation models,'' \emph{arXiv preprint arXiv:2401.00544}, 2023.

\bibitem{siragusa2023conditioning}
I.~Siragusa and R.~Pirrone, ``Conditioning chat-gpt for information retrieval: The unipa-gpt case study,'' in \emph{Proceedings of the Seventh Workshop on Natural Language for Artificial Intelligence (NL4AI 2023) co-located with 22th International Conference of the Italian Association for Artificial Intelligence (AI* IA 2023)}, 2023.

\bibitem{kulkarni2024reinforcement}
M.~Kulkarni, P.~Tangarajan, K.~Kim, and A.~Trivedi, ``Reinforcement learning for optimizing rag for domain chatbots,'' \emph{arXiv preprint arXiv:2401.06800}, 2024.

\bibitem{campese2023quadro}
S.~Campese, I.~Lauriola, and A.~Moschitti, ``Quadro: Dataset and models for question-answer database retrieval,'' \emph{arXiv preprint arXiv:2304.01003}, 2023.

\bibitem{muennighoff2022mteb}
N.~Muennighoff, N.~Tazi, L.~Magne, and N.~Reimers, ``Mteb: Massive text embedding benchmark,'' \emph{arXiv preprint arXiv:2210.07316}, 2022.

\bibitem{zhang2019bertscore}
T.~Zhang, V.~Kishore, F.~Wu, K.~Q. Weinberger, and Y.~Artzi, ``Bertscore: Evaluating text generation with bert,'' \emph{arXiv preprint arXiv:1904.09675}, 2019.

\bibitem{es2023ragas}
S.~Es, J.~James, L.~Espinosa-Anke, and S.~Schockaert, ``Ragas: Automated evaluation of retrieval augmented generation,'' \emph{arXiv preprint arXiv:2309.15217}, 2023.

\bibitem{s_2024}
S.~Kumar, A.~Solanki, and N.~Z. Jhanjhi, ``Rouge-ss: A new rouge variant for the evaluation of text summarization,'' 2024.

\bibitem{banerjee2005meteor}
S.~Banerjee and A.~Lavie, ``Meteor: An automatic metric for mt evaluation with improved correlation with human judgments,'' in \emph{Proceedings of the acl workshop on intrinsic and extrinsic evaluation measures for machine translation and/or summarization}, 2005, pp. 65--72.

\bibitem{barbella2022rouge}
M.~Barbella and G.~Tortora, ``Rouge metric evaluation for text summarization techniques,'' \emph{Available at SSRN 4120317}, 2022.

\end{thebibliography}

\end{document}